# Conceptual Analysis of Lexical Taxonomies: The Case of WordNet Top-Level

Aldo Gangemi[(1)], Nicola Guarino[(2)], Alessandro Oltramari[(2)]
(1) ITBM-CNR, Rome, Italy
(2) LADSEB-CNR, Padova, Italy
aldo@saussure.irmkant.rm.cnr.it,
{Nicola.Guarino, Alessandro.Oltramari}@ladseb.pd.cnr.it

**Abstract**   In this paper we propose an analysis and an upgrade of *WordNet's* top-level synset taxonomy. We briefly review WordNet and identify its main semantic limitations. Some principles from a forthcoming *OntoClean* methodology are applied to the ontological analysis of WordNet. A revised top-level taxonomy is proposed, which is meant to be more conceptually rigorous, cognitively transparent, and efficiently exploitable in several applications.

**Categories & Descriptors**   H3.1 [Information Storage and Retrieval]: Content Analysis and Indexing —*Indexing methods, Linguistic processing, Thesauruses*.

**General Terms**   Experimentation, Theory

**Keywords**   WordNet, ontology, taxonomies, top-level

## 1. Introduction

The main goal of this paper is to present a conceptual analysis of WordNet's top level. We shall use, as far as possible, Guarino and Welty's methodological approach, a powerful (yet not completed) set of theoretical tools for the ontological refinement of taxonomies [4,5]. We shall integrate these tools with techniques derived from ONIONS project, whose original task was the development of a large-scale, axiomatized ontology library for medical terminology. We intend to merge these two methodological patterns into a common methodology called "Onto-Clean".

Presenting here such a complete methodology would be premature. Rather, we intend to show how this methodology can be applied to a practical example, testing at the same time its limits and capabilities.

### 1.1 The WordNet project

Christiane Fellbaum, a member of Princeton team that realized WordNet, describes it as

[ ] a semantic dictionary that was designed as a network, partly because representing words and co n-cepts as an interrelated system seems to be consistent with evidence for the way speakers organize their mental lexicons ([1], p.7).

By exploiting WordNet's structure, users can "build" a personalized cognitive route starting from single English words. Each word may have different senses, shown in the WordNet browser by numbers which identify a definite *synset*, composed by synonyms terms (i.e. <life form, organism, being, living thing>). In this way, not only the gloss corresponding to a certain word sense (as in conventional dictionaries) is made clear, but also the semantic relations to whom the gloss takes part.

The idea of representing world knowledge through a "semantic network" (whose nodes are synsets, and whose arcs are fundamental semantic relations[1]) has been characterizing WordNet's development since 1985. Over the years, there has been an increasing size of the lexicon (version 1.6 contains 90,000 synsets), and a substantial improvement of the entire WordNet architecture, aimed at facilitating a hierarchical arrangement. Princeton researchers have singled out twenty-five classes for nouns, afterwards collapsed into nine (taking into account further psycho-linguistic motivations), corresponding to the "top-concepts" (Unique Beginners) of an ontology. The work described here fits naturally in a productive tradition of research about WordNet (Sensus, CoreLex, EuroWordNet, Simple, FrameNet)

## 2. The OntoClean Methodology: Basic Distinctions

The main point of the OntoClean methodology is the characterization of ontological categories in terms of formal meta-properties. Since these meta-properties are largely independent from a particular ontological commitment, and they are defined for an arbitrary domain, they are a good way to make ontological choices explicit.

The set of meta-properties used in this paper is informally summarized below. Since these notions have been formally defined elsewhere (and the details of this formalization are still subject to change ), we shall limit ourselves to an intuitive description, pointing to [4] and [5] for the technical details. A set of categories (partially) characterized by these meta-properties, corresponding to our preliminary top-level choices, will be described in Section 4.

The term "meta-property" adopted here is based on a fundamental distinction within the domain of discourse: *individuals* or *particulars* on one hand, and *concepts* or *universals* on the other hand. Meta-level properties induce distinctions among concepts, while object-level properties induce distinctions among individuals.

A clear distinction between the object domain and the meta-level domain provides a primary cleaning tool to conceptual modelers: as we shall see in Section 3.2, mixing concepts and individuals in the same taxonomic structure (collapsing IS-A and INSTANCE-OF into a single hyperonimy relation) is a first source of confusion in WordNet.

### 2.1 Formal properties

#### 2.1.1. Rigidity

A property is *essential* to an individual iff it necessarily holds for that individual.

A property is rigid (+R) iff, necessarily, it is essential to all its instances. A property is non-rigid (-R) iff it is not essential to some of its instances, and anti-rigid (~R) iff it is not essential to all its instances.

For example, *person* is usually considered as rigid, since every person is essentially such, while *student* is usually considered as anti-rigid, since every student can posssibly be a non-student.

#### 2.1.2. Identity

A property *carries* an *identity criterion* (+I) iff all its instances can be (re)identified by means of a suitable sameness relation. A property *supplies* an identity criterion iff such criterion is not inherited by any subsuming property.

---
[1] The most important is just synonymy; WordNet also utilizes hyperonymy, meronymy with its variants, antonymy, similarity, troponymy, causation, proper inclusion, etc.

For example, *person* is usually considered as supplying an identity criterion (although determining which one could be hard), while *student* just inherits the identity criterion of *person*, without supplying any further identity criteria.

### 2.1.3. Dependence

An individual *x* is *constantly dependent* on *y* iff, at any time, *x* can't be present unless *y* is fully present, and *y* is not part of *x*. For example, a hole is constantly dependent on (at least part of) its host, and an event is constantly dependent on (at least some of) its participants.

A property P is *constantly dependent* (+D) iff, for all its instances, there exists something they are constantly dependent on.

A property P is *notionally dependent* (+ND) on another property $Q$ iff, whenever $P(x)$ holds, then $Q(y)$ must hold for a different *y* (which is not a proper part of *x*). For instance, *student* can be seen as notionally dependent on *teacher*.

In the following, we shall take "dependent" as synonymous of "constantly dependent", unless otherwise specified.

### 2.1.4. Types and roles

A rigid property that supplies an identity criterion and is not notionally dependent is called a *type*. An anti-rigid property that is notionally dependent is called a *role*. It is a *material role* if it carries (but not supplies) an identity criterion, and a *formal role* otherwise. In our example, *person* would be a type, and *student* a material role. *Part* is an example of formal role, since it carries no identity and is notionally dependent.

Type and Role are examples of formal meta-categories defined by means of multiple meta-properties.

### 2.1.5. Extensionality

An individual is said to be *extensional* iff, necessarily, everything that has the same proper parts is identical to it. A property is extensional (+E) iff, necessarily, all its instances are extensional. It is *anti-extensional* (~E) iff, necessarily, all its instances are non-extensional, so that they can possibly change some parts while keeping their identity.

### 2.1.6. Concreteness

This meta-property is a bit less formal than the previous ones, in the sense that it makes an ontological commitment towards the existence of physical (spatial, temporal or spatio-temporal) locations. We see physical locations as primitive qualities individuals can have (see section 4.5). Without entering into any detail, we shall stipulate that an individual is *concrete* iff it has a physical location. A property whose instances are necessarily concrete will be marked with the meta-property +C (of course, this does not mean that the property itself has a physical location!). Note that an individual can be concrete without being necessarily *real*, or *actual*: Peter Pan is not real but is concrete, since (in some possible world) it must have a physical location.

### 2.1.7. Unity, singularity, and plurality

An individual is *unified* by a certain (suitably constrained) relation *R* iff it is a mereological sum of entities that are bound together by *R*. For instance, the relation *having the same boss* may unify a group of employees in a company. An individual *w* is a *whole* under *R* iff it is maximally unified by *R*, in the sense that *R* is internal to *w*, and no part of *w* is linked by *R* to something that is not part or *w*. For instance, the mereological sum of all the employees in a company forms a whole. An individual is an *essential whole* iff it is necessarily a whole.

A property *P* is said to *carry unity* (+U) if there is a *common* unifying relation *R* such that all the instances of *P* are essential wholes under *R*. A property carries *anti-unity* (~U) if all its instances can possibly be non-wholes. If every instance of *P* is an essential whole, but there is no unifying relation common to all instances of *P*, then we mark *P* with the property *U.

An individual is a *singular* whole iff its unifying relation has a specific topological nature, i.e., more exactly, it is the transitive closure of the relation "strong connection", like that existing between two 3D regions that have a surface in common (again, we are commiting to some notion of space here). The idea is that topological wholes of this kind have a special cognitive relevance, which accounts for the natural language distinction between singular and plural. We shall use strong connection to model intimate material connection, and weak connection (like that existing between regions that only have lines or points in common) to model material contact.

A *plural* individual will be a sum of singular wholes which is not itself a singular whole. Plural individuals may be wholes themselves or not. In the former case they will be called *collections*; in the latter case *pluralities*.

Let us make an example. A piece of coal is an example of a singular whole. A lump of coal will still be a topological whole, but not a singular whole, since the pieces of coal merely touch each other, with no material connection. It will be therefore a plural whole.

## 2.2 Taxonomic constraints imposed by formal properties

As discussed in [4], most of the formal distinctions introduced above impose important constraints on taxonomic relationships. In practice, if a property holds necessarily for all the instances of a certain concept, of course its negation cannot hold necessarily for all the instances of a subsumed concept. This means that, if *F* is a certain formal property, anti-*F* cannot subsume *F*: anti-rigidity cannot subsume rigidity, anti-unity cannot subsume unity, and anti-extensionality cannot subsume extensionality. Indeed, one of the main advantages of the OntoClean methodology is that, after suitably labeling every concept in a taxonomy with its formal properties, we can easily check its ontological consistency, and restructure the taxonomy if necessary.

## 3. WordNet's Preliminary Analysis

## 3.1 Experiment Setting

We applied our methodological principles and techniques to the noun synsets taxonomy of WordNet 1.6. To perform our investigation, we had to adopt some preliminary assumptions in order to convert WordNet's databases[2] into a workable knowledge base. At the beginning, we assumed that the hyponymy relation could be simply mapped onto the subsumption relation, and that the synset notion could be mapped into the notion of concept[3]. Both subsumption and concept have the usual description logics semantics [8]. In order to work with named concepts, we normalized the way synsets are referred to lexemes in WordNet, thus obtaining one distinct name for each synset: if a synset had a unique noun phrase, this was used as concept name; if that noun phrase was polysemous, the concept name was numbered (e.g. window_1). If a synset had more than one synonymous noun phrase, the concept name linked them together with a dummy character (e.g. Equine$Equid).

Firstly, we created a Loom[4] knowledge base, containing, for each named concept, its direct super-concept(s), some annotations describing the quasi-synonyms, the gloss and the synset topic partition, and its original numeric identifier in WordNet; for example:

---

[2] We used the Prolog WordNet database, the Grind database, and some others from the official distribution.
[3] We will show that this assumption is incorrect, and to maintain it an adaptation of WordNet's synset organization is required.
[4] Loom is a knowledge representation system that implements a quite expressive description logic [7].

```
(defconcept Horse$Equus_Caballus
  :is-primitive Equine$Equid
  :annotations ((topic animals)
    (WORD |horse|)
    (WORD |Equus caballus|)
    (DOCUMENTATION "solid-hoofed herbivorous quadruped domesticated since prehistoric times"))
  :identifier |101875414|)
```

| Table 1 - Elements processed in the Loom WordNet kb | |
|---|---|
| noun entries | 116364 |
| equivalence classes: synonyms, spelling variants, quasi-synonyms | 50337 |
| noun synsets (with a gloss and an identifier for each one) | 66027 |
| nouns | 95135 |
| monosemous nouns | 82568 |
| polysemous nouns | 12567 |
| one-word nouns | 70108 |
| noun phrases | 25027 |

The elements processed in the Loom WordNet knowledge base are reported in Table 1. We report in Table 2 an overview of WordNet's noun top-level as translated in our Loom knowledge base. The nine Unique Beginners are shown in boldface.[5]

### 3.2 Main problems found

Let us discuss now the main drawbacks we found after applying the OntoClean methodology to WordNet's conceptual structure.

#### 3.2.1. Confusion between concepts and individuals

The first critical point is the confusion between concepts and individuals. For instance, if we look at the hyponyms of the Unique Beginner Event, we'll find the synset Fall - an individual - whose gloss is "the lapse of mankind into sinfulness because of the sin of Adam and Eve", together with conceptual hyponyms such as Social_Event, Happening, and Miracle.[6] Under Territorial_Dominion we find Macao and Palestine together with Trust_Territory. The latter synset, defined as "a dependent country, administered by a country under the supervision of United Nations", denotes a general kind of country, rather than a specific country as those preceding it. We found many other examples of this sort.

We face here a general problem: the concept/individual confusion is nothing but the product of an "expressivity lack". In fact, if there was an INSTANCE-OF relation, we could distinguish between a concept-to-concept relation (subsumption) and an individual-to-concept one (instantiation). Taking the previous example, we could therefore say that Palestine is an instance of Territorial_Dominion, while Trust_Territory is subsumed by it.

#### 3.2.2. Confusion between object-level and meta-level: the case of Abstraction

The synset Abstraction_1 seems to include both abstract entities, such as Set, Time, and Space, and abstractions (meta-level concepts) such as Attribute, Relation, Quantity. From what the cor-responding gloss expresses, an abstraction "is a general concept formed by extracting common features from specific examples" (New York skyscrapers, Solar System

```
Abstraction_1                              Film
  Attribute                                  Part$Portion
    Color                                      Body_Part
      Chromatic_Color                          Substance$Matter
  Measure$Quantity$Amount$Quantum                Body_Substance
  Relation_1                                     Chemical_Element
  Set_5                                          Food$Nutrient
  Space_1                                    Part$Piece
  Time_1                                     Subject$Content$Depicted_Object
Act$Human_Action$Human_Activity            Event_1
  Action_1                                   Fall_3
  Activity_1                                 Happening$Occurrence$Natural_Event
  Forfeit$Forfeiture$Sacrifice                 Case$Instance
Entity$Something                             Time$Clip
  Anticipation                               Might-Have-Been
  Causal_Agent$Cause$Causal_Agency         Group$Grouping
  Cell_1                                     Arrangement_2
  Inessential$Nonessential                   Biological_Group
  Life_Form$Organism$Being$                  Citizenry$People
  Object$Physical_Object                   Phenomenon_1
    Artifact$Artefact                        Consequence$Effect$Outcome
      Edge_3                                 Levitation
      Skin_4                                 Luck$Fortune
      Opening_3                            Possession_1
      Excavation$                            Asset
      Building_Material                      Liability$Financial_Obligation$
        Mass_5                               Territory$Dominion$
        Cement_2                             Transferred_Property$
        Bricks_and_Mortar                  Psychological_Feature
        Lath_and_Plaster                     Cognition$Knowledge
    Body_Of_Water$Water                        Structure
    Land$Dry_Land$Earth$                     Feeling_1
    Location                                 Motivation$Motive$Need
    Natural_Object                         State_1
      Blackbody_Full_Radiator                Action$Activity$Activeness
      Body_5                                 Being$Beingness$Existence
      Universe$Existence$Nature$             Condition
      Paring$Parings                         Damnation$Eternal_Damnation
```

---

[5] It must be noticed that the sense numeration reported in our Loom kb is different from the WordNet's original one. Nevertheless, the reader will easily recognize the synsets we are referring to.

[6] In the text body, we usually do not report all the synonyms of a synset (or their numeration), but only the most significative ones.

**Figure 1. WordNet's Top Level.**

planets, Italian ministers). Abstraction seems to be intended therefore as a psychological process of generalization, in accordance to Locke's notion of Abstraction ([6], p.211). This meaning seems to fit the latter group of terms (Attribute, Relation, Quantity), but not to the former, which would be considered as abstract under a different notion of abstraction, namely not being extended in space/time. Moreover, attributes, relations, and quantities appear to be meta-level concepts, while Set, Time, and Space seem to belong to the object domain.

### 3.2.3. Formal properties violations in subsumption relation

Moving now to the field of meta-level categories, the most common violation we have registered is about rigidity, which is bound to the distinction between roles and types. A role cannot subsume a type. Let's see an important clarifying example.

In its first sense, Person (which we consider as a type) is subsumed by two different concepts, Organism and Causal_Agent. Organism can be conceived as a type, while Causal_Agent as a formal role. The first subsumption relationship is feasible, while the second one shows a rigidity violation. We propose therefore to drop it.

Someone could argue that every person is necessarily a causal agent, since "agentivity" (capability of performing actions) is an essential property of human beings. In this case Causal_Agent would be intended as a synonym of "intentional agent", and would be considered as rigid. But in this case it would have only hyponyms denoting things that are (essentially) causal agents, including animals, spiritual beings, the personified Fate, and so on. Unfortunately, this is not what happens in WordNet: Agent, one of Causal_Agent hyponyms, is defined as: "an active and efficient cause; capable of producing a certain effect; (the research uncovered new disease agents)". Causal_Agent subsumes roles such as Germicide, Vasoconstrictor, Antifungal. Instances of these concepts are not causal agents essentially. This means that considering Causal_Agent as rigid would introduce further inconsistencies.

These considerations allow us to add a pragmatic guideline to our methodological techniques: when deciding about the formal meta-property to attach to a certain concept, it is useful to look at all its children.

### 3.2.4. Heterogeneous levels of generality

Going down the different layers of WordNet's top level, we register a certain "heterogeneity" in their intuitive level of generality. For example, among the hyponyms of Entity there are types such as Physical_Object, and roles such as Subject. The latter is defined as "something (a person or object or scene) selected by an artist or photographer for graphic representation", and has no hyponyms (indeed, almost any entity can be an instance of Subject, but none is necessarily a subject)[7].

For Animal (subsumed by Life_Form) this heterogeneity becomes clearer. Together with classes such as Chordate, Larva, Fictional_Animal, etc., we find out more specific concepts, such as Work_Animal, Domestic_Animal, Mate_3, Captive, Prey, etc. We are induced to consider the formers as types, while the latters as roles.

In other words, we discovered that, if at a first sight some synsets sound intuitively too specific when compared to their siblings, from a formal point of view we may often explain their "different generality" by means of the distinction between types and roles.

## 4. The OntoClean preliminary top-level

Let us now introduce the top-level categories that we have used for our experiment. They represent a first draft of the top-level distinctions we plan to use in our OntoClean methodology. They have been chosen in order to be as general and neutral as possible, although (differently from the formal properties used to characterize them) they reflect our cognitive bias, aimed at capturing the ontological categories lying behind natural language and human commonsense.

For the time being, we have also avoided to impose a strong taxonomic structure to these categories, focusing on producing a more or less flat list of basic concepts. In the future, a further restructuring will be probably needed (for instance to better account for the *continuants/occurrents* distinction, which we have only marginally addressed here). The nature of the categories described below would not change, however.

All categories listed below are considered to be *rigid*, as they are assumed to reflect essential properties of their instances. It is this ascription of essentiality which is the result of our commonsense conceptualization: when an amount of matter is shaped to form an object, the deci-sion to consider this "objecthood" as an essential property is just the result of our conceptualization, since of course there is no ontological need for that.

---

[7] We can draw similar observations for relation_1 and set_5 with respect to abstraction_1, etc.

## 4.1 Aggregates (~D, ~U)

The common trait of aggregates is that they are independent entities, and none of them is an essential whole. This means that the corresponding property carries anti-unity (~U). We consider two kinds of aggregates: *Amounts of matter* and *Pluralities*. The latter can be also called *groups*, or perhaps *sets*; we prefer however to use *set* for abstract entities, and *group* does sometimes denote something with an intrinsic unity. Pluralities are just *mere sums of wholes* which are not themselves essential wholes. Amounts of matter are extensional (+E), in the sense that they change their identity when they change some parts; mere pluralities can be considered as *pseudo-extensional*, in the sense that they change their identity when a member is changed, while a change in the parts of a member may be allowed.

## 4.2 Objects (~D, *U)

The main characteristic of objects is that all of them are independent essential wholes. This does *not* mean that the corresponding property (*being an object*) carries +U, since there is no *common* unity criterion for objects. Among objects, we distinguish between *physical bodies* and *ordinary objects*. Bodies are considered to be extensional (+E), while ordinary objects are not (~E). This means that we assume that all ordinary objects can change some of their parts while keeping their identity.

## 4.3 Events (+D, +E)

Events are *things that happen in time*, i.e. *temporal occurrences*. They are assumed to be dependent on those objects that are their *participants*. Therefore, an object that participates to an event is not part of that event. Events can have temporal parts, like the first movement of a symphony, or spatial parts, like the strings playing within that symphony. The important point is that all parts of an event are *essential parts*: if an event would change any of its parts, it would be a different event. Events are therefore extensional. Our work on events is pretty much in-progress now, so we are not currently in the position to further elaborate on them. We only point out the necessity to distinguish objects from events is just a result of our cognitive-linguistic bias, and we do not make any metaphysical commitment regarding the "primacy" of objects on events.

## 4.4 Features (+D, -E, *U)

Features are "parasitic entities", that exist insofar their *host* exists. Typical examples of features are holes, bumps, boundaries, or spots of color. Features may be *relevant parts* of their host, like a bump or an edge, or *dependent regions* like a hole in a piece of cheese, the underneath of a table, the front of a house, or the shadow of a tree, which are not parts of their host. All features are essential wholes, but no common unity criterion may exist for all of them. However, typical features have a topological unity, as they are *singular* entities.

## 4.5 Qualities (+D, +E, +U)

Qualities and properties are often considered as synonymous, but they are not. Take a particular object, like a rose: depending on its nature (and our way of perceiving it) it will exibit some individual qualities, like a specific color, a size, a smell, etc. The way we classify these qualities may depend on our conceptualization of them, which strongly depends on our culture and perceptual capabilities. Properties like *red*, *big*, *sweet* are the result of classifying each of these qualities with respect to a specific *conceptual space* [3]: so the rose is red because its color is located in a certain region in the colors conceptual space. When we say that "red is a color" we are talking of a region in this space; when we say "I like the color of this rose" we are talking of an individual quality. We assume that individual qualities do not change in time, while their *conceptual location* in a conceptual space can vary in time. The main characteristics of qualities is therefore their being located in conceptual spaces, whose topological structure depends on the quality being considered.

Speaking of their meta-properties, qualities are dependent entities. We also assume that they have no proper parts, so that they are trivially extensional and are trivially wholes. Finally, we assume that qualities of objects are physically located where the objects are located, so that qualities of concrete objects are themselves concrete.

An important remark is that we take spatial and temporal locations of objects as individual qualities, too. This means that geometric space and time are considered as conceptual spaces.

### 4.6 Abstractions (~C)

Abstractions are entities that are not concrete, that is, they do not have a physical location. Conceptual spaces are the first example of abstractions: time, geometric space, length, color are all conceptual spaces, with different topological structure. Terms like *red, long, old, recent* correspond to *regions* in a conceptual space (and therefore to particulars, not universals). We can describe therefore the structure of a conceptual space with a first-order theory, using topological notions: for instance, we can say that *red* is adjacent to *brown*. Depending on the way a conceptual space is partitioned, and on the *metric* (if any) imposed on it, we can have different equivalent ways of describing a certain quality.

Other abstractions are *sets*, *symbols, propositions, structures.*

## 5. WordNet Cleaned up: mapping WordNet into the OntoClean top-level

Let us consider now the results of integrating the WordNet top concepts into our top-level. According to the OntoClean methodology, we have concentrated first on the so-called *backbone taxonomy*, which only includes the rigid properties. Formal and material roles have been therefore excluded from this preliminary work.

While comparing WordNet's unique beginners with our ontological categories, an extreme heterogeneity appears evident: for example, Entity looks like a "catch-all" class containing concepts hardly classifiable elsewhere, like Anticipation, Imaginary_Place, Inessential, etc. These synsets have hardly a few children, and have been excluded by our analysis by now.

The results of our integration work are sketched in Table 2. Our categories are reported in the first column; the second column shows the WordNet synsets that are *covered* by such categories (i.e., they are either equivalent to or included by them); the third column shows some hyponims of these synsets that were rejected according to our methodology. Finally, the last column shows further hyponyms that have been appended under our categories, coming from different places in WordNet. The problems encountered for each category are discussed below.

### 5.1 Aggregate, Object, Feature

ENTITY$SOMETHING is a very confused synset. As sketched in the table, a lot of its hyponyms have to be "rejected": in fact there are roles (Causal_Agent, Subject_4), unclear synsets (Location[8]) and so on. This Unique Beginner maps partly to our Aggregate and partly to our Object category. Some hyponyms of Physical_Object are mapped to our new top concept Feature.

By removing roles like Arrangement and Straggle, GROUP$GROUPING becomes a partition of the Ordinary Object category (namely the third child of the top concept Object.

---

[8] Referring to Location, we find roles (There, Here, Home, Base, Whereabouts), instances (Earth), and geometric concepts like Line, Point, etc.).

In fact, hyponyms like Collection, Social_Group, Biological_Group ecc., are nothing but plural objects, supporting a clear unity criterion.

POSSESSION_1 is a role, and it includes both roles and types. In our opinion, the synsets marked as types (Asset, Liability, etc.) should be moved towards lower levels of the ontology,

### Table 2

| Top Categories | Covered Synsets | Rejected Hyponyms | Imported Hyponyms |
|---|---|---|---|
| *Aggregate* | Aggregate_2 ! | | |
| Amount_of_matter | Substance$Matter* | Bedding_Material, Ballast, Atom, Philosopher's_ Stone | Mass_5, Cement_2, Substance, |
| Plurality | | | |
| *Object* | Entity$Something* | Anticipation, Causal_Agent, Imaginary_Place, Substance | |
| Physical_Body | Natural_Object* | Dead_Body, Constellation, Stone, Nest | |
| Ordinary_Object | Physical_Object* Group | Finding, Catch, Vagabond; arrangement, | |
| *Feature* | | | |
| Relevant_part | part$portion* fragment | | Edge_3, Skin_4, Paring$Parings, |
| Plural_Feature | | | |
| Dependent_Region | | | Opening_3, Excavation$hole_in_ the_Ground, |
| *Quality* | Attribute* | Trait, Ethos, Inheritance | |
|    Time | time_interval$interval* | Eternity, Greenwich_Mean_Time, Present, Past, Future | |
|    Color | chromatic_color | | |
| *Abstraction* | | | |
| Abstract_Entity | | | Statement_1, Cognition, Arrangement_2, Ownership_1, |
|    Proposition | Proposition_1 | | |
|    Set | set_5 | | |
| Quality_Space | Attribute* | Trait, Ethos, Inheritance | |
|    Space | space_1 | Subspace, | |
|    Time | time_interval$interval* | Eternity, Greenwich_Mean_Time, Present, Past, Future | |
|    Color | chromatic_color | | |
| *Event* | | | PHENOMENON_1 *, STATE_1*, Cognitive Event, EVENT_1*, ACT* |

***TABLE 2 LEGEND***: Hyponyms marked with "*" are heterogeneous (some of them are to be moved elsewhere, some are roles, or some are instances); those marked with "!" have no hyponyms; those in upper case are WordNet Unique Beginners; those in italic are Top Concepts of the OntoClean ontology.

since their meanings seem to deal more with a specific domain - the economic one - than with a set of general concepts (except some concepts that can be mapped to Abstraction). This means that the remainder branch is also to be eliminated from the top level, because of its overall anti-rigidity (the peculiarity of roles).

## 5.2 Abstraction, Quality

ABSTRACTION_1 is the most heterogeneous Unique Beginner. It contains abstracts (Set_5), quality spaces (Chromatic_Color), qualities (mostly from the synset Attribute) and a hybrid concept (Relation_1) that contains abstracts, other entities, and even meta-level categories. Each child synset has been mapped appropriately. As we can see from the table, our Abstraction top concept relates to its WordNet homonymous only with regard to few hyponyms.

PSYCHOLOGICAL FEATURE contains both abstract entities (Cognition) and Events (Feeling_1), and its children synsets have been mapped accordingly (we have created a new Cognitive_Event concept to support such synsets under Event).

## 5.3 Event

EVENT_1, PHENOMENON_1, STATE_1, ACT are globally mapped to our Event category, although — by simply looking at their children — it seems quite hard to explicit any criteria to maintain the original distinctions. A comprehensive analysis of lower taxonomic levels is needed, based for instance on linguistics literature, which applies such criteria as durativeness, dynamicity, intentionality, completion, etc. (cf. [2]), and also on a deep analysis of recent top ontologies as those developed for SIMPLE and EuroWordNet projects. A formal characterization of these criteria is an ongoing work.

## Conclusions

The final results of our integration effort are sketched in Figure 2. This can be seen as a preliminary restructuring of WordNet's top level by means of the OntoClean methodology. Our results show that a serious ontological improvement is needed in order to thoroughly exploit the semantic richness of WordNet. For this purpose, exploiting the ontological constraints resulting from the OntoClean methodology is a big advantage. Our research is still in progress, and more experiments are needed to improve the methodology itself: we hope we have paved the way for future work.

## Acknowledgements

We would like to thank Claudio Masolo, Chris Partridge, Domenico Pisanelli, and Geri Steve for the fruitful discussions and comments on the earlier version of this paper. This work was partly supported by the Eureka Project IKF (E!2235, Intelligent Knowledge Fusion), and the National project TICCA (Tecnologie Cognitive per l'Interazione e la Cooperazione con Agenti Artificiali).

*LADSEB-CNR and ITBM-CNR publications are retrievable on-line at the following URLs:*

http://www.ladseb.pd.cnr.it/infor/ontology/ontology.html

http://saussure.irmkant.rm.cnr.it

```
Aggregate                                   Dependent_Region
    Amount of matter                            opening_3
        body_substance                          excavation$hole_in_the_ground
        chemical_element
        mixture
        compound$chemical_compound         Quality
        mass_5                                  position$place
        fluid_1                                 time_interval$interval*
    Plurality                                   chromatic_color

Object                                      Abstraction
    Physical_Body                               Abstract_Entity
        blackbody$full_radiator                     cognition$knowledge
        body_5                                      structure
        universe$existence$nature$creation
                                                    statement_1
    Ordinary _Object                                    proposition
        collection$aggregation                      symbol
        biological_group                            set_5
        social_group
        kingdom                                 Quality_Space
                                                    space_1
        geographical_object                         time_1
            Body_Of_Water$Water                     time_interval$interval*
            Land$Dry_Land$Earth$                    chromatic_color
        body$organic_structure
        artifact$artefact*
    Life_Form$Organism$Being$               Event
                                                Event_1
Feature                                         Phenomenon_1
    Relevant_Part                               State_1
        edge_3                                  Act$Human_Activity
        skin_4                                  Cognitive_event
        paring$parings
```

**Figure 2. WordNet cleaned up: mapping WordNet into the OntoClean top-level**